# GA-PSO-Optimized Neural-Based Control Scheme for Adaptive Congestion Control to Improve Performance in Multimedia Applications

Mansour Sheikhan[1], Ehasn Hemmati[2], Reza Shahnazi[3]
1- Department of Communication Engineering, South Tehran Branch, Islamic Azad University, Tehran, Iran.
Email: msheikhn@azad.ac.ir
2- Department of Electronic Engineering, South Tehran Branch, Islamic Azad University, Tehran, Iran.
Email: st_e_hemmati@azad.ac.ir
3- Modeling and Optimization Research Center in Science and Engineering, South Tehran Branch, Tehran, Islamic Azad University, Iran.
Email: shahnazi@ieee.org

**ABSTRACT:**
Active queue control aims to improve the overall communication network throughput, while providing lower delay and small packet loss rate. The basic idea is to actively trigger packet dropping (or marking provided by explicit congestion notification (ECN)) before buffer overflow. In this paper, two artificial neural networks (ANN)-based control schemes are proposed for adaptive queue control in TCP communication networks. The structure of these controllers is optimized using genetic algorithm (GA) and the output weights of ANNs are optimized using particle swarm optimization (PSO) algorithm. The controllers are radial bias function (RBF)-based, but to improve the robustness of RBF controller, an error-integral term is added to RBF equation in the second scheme. Experimental results show that GA- PSO-optimized improved RBF (I-RBF) model controls network congestion effectively in terms of link utilization with a low packet loss rate and outperforms Drop Tail, proportional-integral (PI), random exponential marking (REM), and adaptive random early detection (ARED) controllers.

**KEYWORDS:** Adaptive Control, Queue, Communication Network, Optimization, GA, PSO, Neural Network.

## 1. INTRODUCTION

To improve the performance of a communication network, in terms of packet loss rate, link utilization and delay, active queue management (AQM) technology has been proposed to replace Drop Tail queue management method [1]. Buffer management for the transmission control protocol (TCP)/Internet Protocol (IP) routers plays an important role in the congestion control. A small buffer generally achieves a low queuing delay, but suffers from excessive packet losses and low link utilization, and vice versa.

Active queue control schemes are those policies of router queue management that allow for the detection of network congestion, the notification of such occurrences to the hosts, and the adoption of a suitable control policy. A simple policy like the widely used first-in-first-out (FIFO) Drop Tail often causes strong correlations among packet losses, resulting in the well-known "TCP synchronization" problem [1]. AQM policies based on control theory consider the intrinsic feedback nature of congestion systems. Sources adjust their transmission rates according to the level of congestion.

In recent years, several approaches have been proposed as AQM policies [2, 3]. AQM policies provide better network utilization and lower end-to-end delays than Drop Tail method. The development of new AQM routers will play a key role in meeting tomorrow's increasing demand and improving performance in voice over IP (VoIP), class of service (CoS), and streaming video applications where the packet size and session duration exhibit significant variations.

Random early detection (RED) algorithm is the earliest and the most prominent of AQM schemes [4] which is recommended by the Internet Engineering Task Force (IETF), to be deployed in the Internet. However, the behavior of RED strongly depends on tuning parameters for every specific case and average queue size varies with the level of congestion.

In addition, some modified schemes of RED including flow random early detection (FRED) [5], adaptive RED (ARED) [6, 7], balanced RED (BRED) [8], stabilized RED (SRED) [9], dynamic RED (DRED) [10], BRED with virtual buffer occupancy (BRED/VBO) [11], exponential RED (ERED) [12], neural network-based RED (NN-RED) [13], loss ratio and rate control RED (LRC-RED) [14], novel autonomous proportional and differential RED (PD-RED) [15, 16], and robust RED (RRED) [17] have been proposed in recent two decades. Furthermore, other algorithms such as adaptive virtual queue (AVQ) [18], random exponentially marking (REM) [19], BLUE [20], GREEN [21] and YELLOW [22] have been proposed as alternatives to RED.

Also, several control strategies have been proposed for active queue control such as classic proportional-integral (PI) [23], proportional-differential (PD) [24], proportional-integral-differential (PID) controllers [25], sliding mode control [26, 27], optimal control [28, 29], virtual rate control (VRC) [30], adaptive control [31], observer-based control [32], fuzzy control [33-36], predictive functional control (PFC) [37], robust control [38], variable structure control (VSC) [39], and neural-based control [13, 40-42]. In addition, several optimization techniques have been proposed to improve the controller performance such as genetic algorithm (GA) [35] and particle swarm optimization (PSO) algorithm [43].

In this paper, two artificial neural networks (ANN)-based control schemes are proposed for adaptive queue control in TCP communication networks. The structure of these controllers is optimized using GA and the output weights of ANNs are optimized using PSO algorithm. Due to the nonlinear characteristic of traffic in the communication network, radial basis function (RBF) neural model is used to control the queue and achieve desired quality of service (QoS). Also, as an improved RBF model, an error-integral term is added to RBF equations to increase the robustness and improve the performance of active queue controller. This improved RBF model is called I-RBF in this paper. Suggestion of a robust neural-based controller with optimized structure and weights for adaptive congestion control is the main contribution of this paper which has not been reported before in the literature.

The rest of paper is outlined as follow. The background of the methods used in the proposed scheme, including GA, PSO and RBF, is reviewed in Section 2. In Section 3, the dynamic model of TCP is introduced. The proposed GA-PSO-optimized RBF and I-RBF neural controllers are illustrated in Sections 4 and 5, respectively. Simulation details and experimental results are given in Section 6. Finally, the work is concluded in Section 7.

## 2. BACKGROUND
### 2.1. Genetic Algorithm
The genetic algorithm is a method for solving optimization problems based on natural selection, the process that drives biological evolution. The genetic algorithm repeatedly modifies a population of individual solutions. At each step, the genetic algorithm selects individuals randomly from the current population to be parents and uses them to produce the children for the next generation. There are several methods for selecting parents such as stochastic uniform selection, remainder selection, roulette selection and tournament selection [44].

In this work, the optimized number of neurons in hidden layer of RBF neural model is selected by GA. The fitness function measures the quality of the solution in GA and is application-dependent. In this application, the fitness function is chosen as follows [45]:

$$F = \frac{1}{(MSE)^2} \quad (1)$$

At the beginning, the generated values by fitness function are not suitable for selection process of patterns. So, fitness scaling is necessary to map those raw values into a new suitable range for the selection function. The range of scaled values affects the performance of genetic algorithm. In this study, "Rank" fitness scaling function is used to remove the effects of raw scores spread. To create the next generation, GA uses elite children that are individuals with the best fitness values in the current generation.

### 2.2. Particle Swarm Optimization Algorithm
PSO is a population based stochastic optimization technique which does not use the gradient of the problem being optimized, so it does not require being differentiable for the optimization problem as is necessary in classic optimization algorithms. Therefore it can also be used in optimization problems that are partially irregular, time variable, and noisy.

In PSO algorithm, each bird, referred to as a "particle", represents a possible solution for the problem. Each particle moves through the $D$-dimensional problem space by updating its velocities with the best solution found by itself (cognitive behavior) and the best solution found by any particle in its neighborhood (social behavior). Particles move in a multidimensional search space and each particle has a velocity and a position as follow:

$$v_i(k+1) = v_i(k) + \gamma_{1i}(P_i - x_i(k)) + \gamma_{2i}(G - x_i(k)) \quad (2)$$

$$x_i(k+1) = x_i(k) + v_i(k+1) \quad (3)$$

where $i$ is the particle index, $k$ is the discrete time index, $v_i$ is the velocity of $i$th particle, $x_i$ is the position of $i$th particle, $P_i$ is the best position found by $i$th particle (personal best), $G$ is the best position found by swarm (global best) and $\gamma_{1,2}$ are random numbers in the interval [0,1] applied to $i$th particle. In our simulations, the following equation is used for velocity [46]:

$$v_i(k+1) = \varphi(k)v_i(k) + \alpha_1\left[\gamma_{1i}(P_i - x_i(k))\right] + \alpha_2\left[\gamma_{2i}(G - x_i(k))\right] \quad (4)$$

in which $\phi$ is inertia function and $\alpha_{1,2}$ are acceleration constants. The flowchart of PSO algorithm is summarized in Fig. 1.

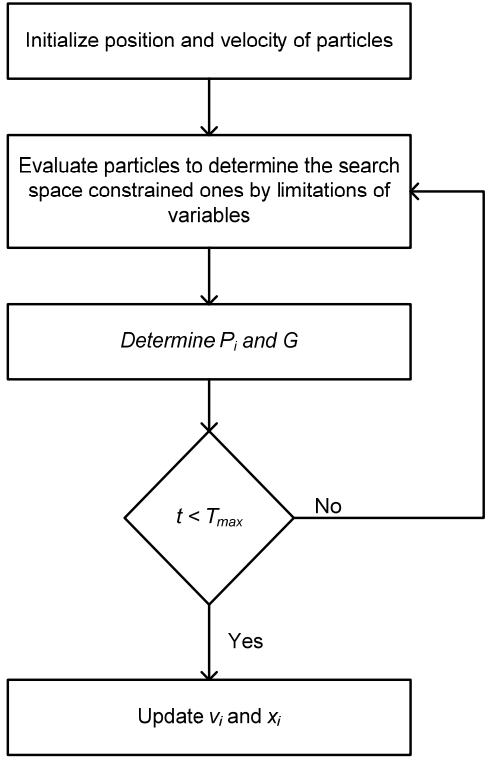

**Fig. 1.** Standard PSO flowchart

### 2.3. RBF neural model

The RBF neural network involves three different layers. The input layer is the set of source nodes and the second layer consists of RBF nodes. The transformation from the input space to the hidden-unit space is nonlinear. On the other hand, the transformation from the hidden space to the output space is linear.

The output $y$, of a Gaussian RBF network is evaluated from the input vector $x$, as follows:

$$y(x) = w^T \exp(\frac{-\|x-c\|^2}{r^2}) \quad (5)$$

where $y(x)$ is approximate function, $w$ is weight for basis function called output weight. $c$ and $r$ are the mean and spread of the Gaussian function, respectively.

### 3. DYNAMIC MODEL OF TCP ACTIVE QUEUE CONTROL

A simplified version of TCP model, which expresses the coupled nonlinear differential equations which reflect the dynamics of TCP with the average TCP window size and the average queue length, is given by (6) and (7) [47]. It is noted that the packet-dropping probability is between 0 and 1, so the following nonlinear time-delayed system with a saturated input are proposed:

$$\frac{dW}{dt} = \frac{1}{R(t)} - \frac{W(t)}{2}\frac{W(t-R(t))}{R(t-R(t))}sat(u(t)) \quad (6)$$

$$\frac{dq}{dt} = \begin{cases} -C + \frac{N(t)}{R(t)}W(t) & \text{if } q(t)>0 \\ \max\left\{0, -C + \frac{N(t)}{R(t)}W(t)\right\} & \text{if } q(t)=0 \end{cases} \quad (7)$$

where $W(t)$ is the mean TCP window size (in packets), $q(t)$ is the queue size (in packets), $R(t)$ is the round trip time (RTT) (in seconds) and equals to $q/C+T_p$, $C$ is the link capacity (in packets/second), $N(t)$ is the number of TCP connections, and $p(t)$ is the packet mark/drop probability.

The saturated input is expressed by the following nonlinearity:

$$sat(u(t)) = \begin{cases} 1 & u(t) \geq 1 \\ u(t) & 0 \leq u(t) < 1 \\ 0 & u(t) < 0 \end{cases} \quad (8)$$

### 4. GA-PSO-OPTIMIZED RBF CONTROLLER

In this section, an RBF controller, which its structure is optimized using GA and its output weights are optimized using PSO algorithm, is proposed to achieve the desired queue length efficiently considering delay effects and a saturated input. The RBF controller generates a control input term, $u(t)$, as mentioned in (6). The output error signal is defined as follows:

$$e(t) = q(t) - q_t \quad (9)$$

where $q_t$ denotes the target queue length. In this way, a Gaussian RBF controller with an input $e(t)$ and an output $u(t)$ is expressed as follows:

$$u(t) = w^T \varphi(e) \quad (10)$$

where $w$ is the weight of hidden layer of RBF network and $\varphi$ is defined by:

$$\varphi_i(e) = \exp(-\frac{(\|e-c_i\|)^2}{\sigma_i^2}) \quad (11)$$

where $c_i$ is the mean and $\sigma_i$ is the spread of $i$th radial bias function.

In order to measure the performance of the proposed closed-loop controller, the integral absolute error (IAE) measure is used with the following equation:

$$IAE = \frac{1}{T}\int_0^T |e(\tau)|d\tau \quad (12)$$

In fact, the IAE depends on the controller parameters which here are the output weights of RBF that its structure is optimized. So, the goal is to find the optimum values of RBF parameters which made the IAE minimum. The proposed GA-PSO-optimized neural-based control scheme is depicted in Fig. 2.

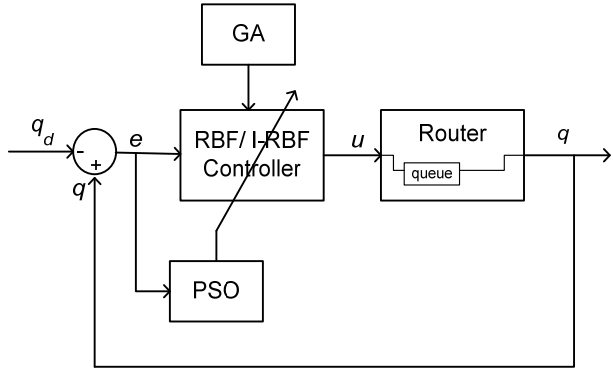

**Fig. 2.** GA-PSO-optimized neural-based scheme for adaptive queue

## 5. GA-PSO-OPTIMIZED I-RBF CONTROLLER

In order to improve the controller performance, an error-integral term is added to the proposed RBF controller. In this scheme, the relation between controller input $e(t)$ and the control signal $u(t)$ is defined as follows:

$$u(t) = w^T \varphi(e) + w_I \int_0^t e(\tau)\,d\tau \qquad (13)$$

where $w_I$ is the integral gain and $w$ is the output weight of RBF network and $\varphi$ is defined as (11).

Again IAE is employed in order to measure the performance of the proposed closed-loop controller. Similarly, if the smaller value of IAE is achieved, then the better controller is designed. Hence, the goal is to find the optimum values of RBF parameters and $w_I$ such that the IAE becomes minimum. PSO algorithm is again used in this scheme to tune the controller parameters, as shown in Fig. 2, since it has good performance in continuous optimization problems.

## 6. SIMULATION AND EXPERIMENTAL RESULTS

The effectiveness of proposed GA-PSO-optimized RBF-based controllers is verified in this section through a series of numerical simulations using ns-2 (Network Simulator-2) tool with the dumbbell network topology (Fig. 3). In dumbbell network topology, the transport agent is based on TCP-Reno, where multiple TCP connections share a single bottleneck link. Each link capacity and corresponding propagation delay is also depicted in Fig. 3.

Before performance evaluation in different scenarios in this section, the parameter setting of the optimized RBF and I-RBF in controller design for TCP active queue control is discussed.

In our simulation of genetic algorithm, the population size is assumed to be 40. Two elite children, 26 crossover children, and 12 mutation children are used. It is noted that the fraction of individuals that is used in crossover process is set to 0.7. The Gaussian function is used as the mutation function. The amount of mutation is decreased at each new generation (proportional to the standard deviation of Gaussian distribution). *Shrink* parameter determines the rate of this decrement. The standard deviation of Gaussian distribution is decreased linearly until its final value reaches to (1–*Shrink*) times of its initial value at the first generation. The value of *Shrink* parameter is set to 1 in our simulations. The optimum number of hidden nodes is achieved as five neurons when the fitness value (as defined in (1)) was $F=49.2\times10^2$.

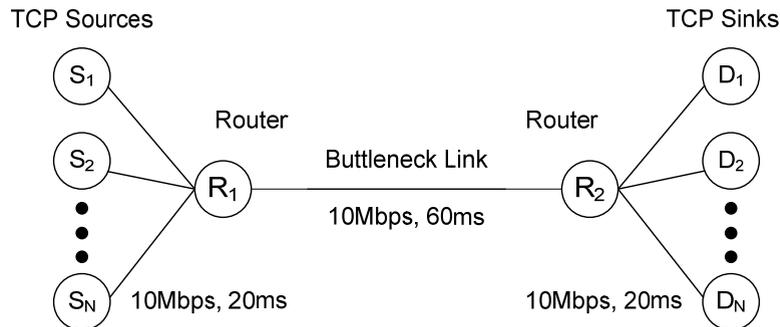

**Fig. 3.** Network topology details for performance evaluations in different scenarios

The initial values of PSO parameters are set according to values depicted in Table 1.

For a TCP network modeled by (6) and (7), it is assumed that $N=100$ homogeneous TCP connections share one bottleneck link with a capacity of 10 Mbps; i.e., $C=1250$ (packets/second). Furthermore, the propagation delay of the bottleneck link capacity is set to $T_p=60$ msec, and the desired queue size is assumed as $q_d=150$ packets. The time duration of queue monitoring, $T$, is set to 100 seconds. The maximum buffer size of each router is assumed to be 300 packets; each packet has a size of 1000 bytes. The spread of all RBFs is set to 40 and the mean is distributed between -150 and 150 steps by 75. Table 2 shows the optimal parameter setting of RBF and I-RBF controllers as the outputs of PSO algorithm with initial condition $q(0) = 0$.

**Table 1.** PSO algorithm parameters setting

| Parameter | Value |
|---|---|
| Size of population | 20 |
| Maximum particle velocity | 4 |
| $\alpha_1$ | 2 |
| $\alpha_2$ | 2 |
| Initial inertia weight | 0.9 |
| Final inertia weight | 0.2 |
| Maximum number of iterations | 300 |

The convergence curve of the IAE value against number of iterations for these controllers is shown in Fig. 4.

To demonstrate the robustness of the proposed queue controllers, the dynamic traffic changes and different RTTs in the simulated network are taken into account in the simulations. The simulation results for these conditions using the proposed control strategies are compared with other AQM schemes such as Drop Tail, PI [47], REM [48], and ARED [7]. The parameter setting of mentioned AQM schemes are listed in Table 3.

The responses of link utilization and packet loss rate for the different number of users and various bottleneck link propagation delays are studied through simulations.

In this case, the number of users is varied between 70 and 160 and various propagation delays between 20 and 140 msec are considered. In Figs. 5 and 6, the link utilization and packet loss rate for different number of users are depicted, respectively.

**Table 3.** Parameter setting of PI, REM and ARED simulated controllers in this study

| AQM scheme | Parameter setting |
|---|---|
| PI [47] | $a = 1.822 \times 10^{-5}$, $b = 1.816 \times 10^{-5}$, $T = 1/160$ s |
| REM [48] | $\gamma = 0.001$, $\varphi = 1.001$ |
| ARED [7] | $min_{th} = 100$, $max_{th} = 215$, $w_g = 1-\exp(-1/C)$ |

As can be seen, the link utilization of proposed GA-PSO-optimized I-RBF controller is higher than other queue controllers except for 160 TCP connections; that is very close to the ARED scheme. Furthermore, for all mentioned queue controllers, the link utilization is raised when the number of users is increased. As shown in Fig. 6, the proposed GA-PSO-optimized I-RBF controller has packet loss rate a little more as compared to the PI and ARED schemes. But the difference between optimized RBF controller and the others is almost noticeable. Link utilization of Drop Tail is almost constant and it is less than all mentioned queue control schemes except REM. On the other hand, the packet loss rate of Drop Tail method is also less as it uses whole queue length.

Figures 7 and 8 depict the responses of link utilization and packet loss rate in different propagation delays, respectively.

As can be seen, REM and Drop Tail link utilizations are almost less than other mentioned queue control schemes for $RTT \leq 100$ msec. The packet loss rate of Drop Tail scheme is also less than other methods. The performance of PI, ARED, RBF and I-RBF is very close to each other and in all of them the link utilization is decreased as the delay increases. However, the packet loss for I-RBF is higher than PI and ARED, but it is very close to them. Since the link utilization is better for it, so it can be neglected in this case.

**Table 2.** PSO-optimized parameters and IAE of proposed adaptive queue controllers

| Adaptive queue control scheme | Optimal weights of proposed controllers | | | | | | IAE |
|---|---|---|---|---|---|---|---|
| | $w_1$ | $w_2$ | $w_3$ | $w_4$ | $w_5$ | $w_I$ | |
| GA-PSO-optimized RBF | -1 | -1 | 0.340 | 0.337 | 1 | Not-Applicable | 0.363 |
| GA-PSO-optimized I-RBF | -1 | -0.961 | 0.345 | 0.994 | 0.998 | $7.0813 \times 10^{-4}$ | 0.383 |

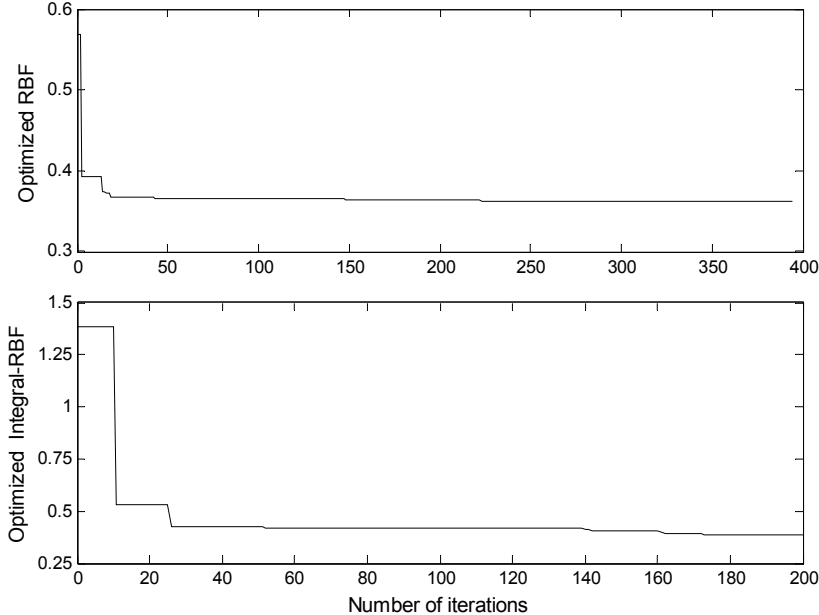

**Fig. 4.** Convergence of IAE values for optimized RBF and I-RBF adaptive queue controllers

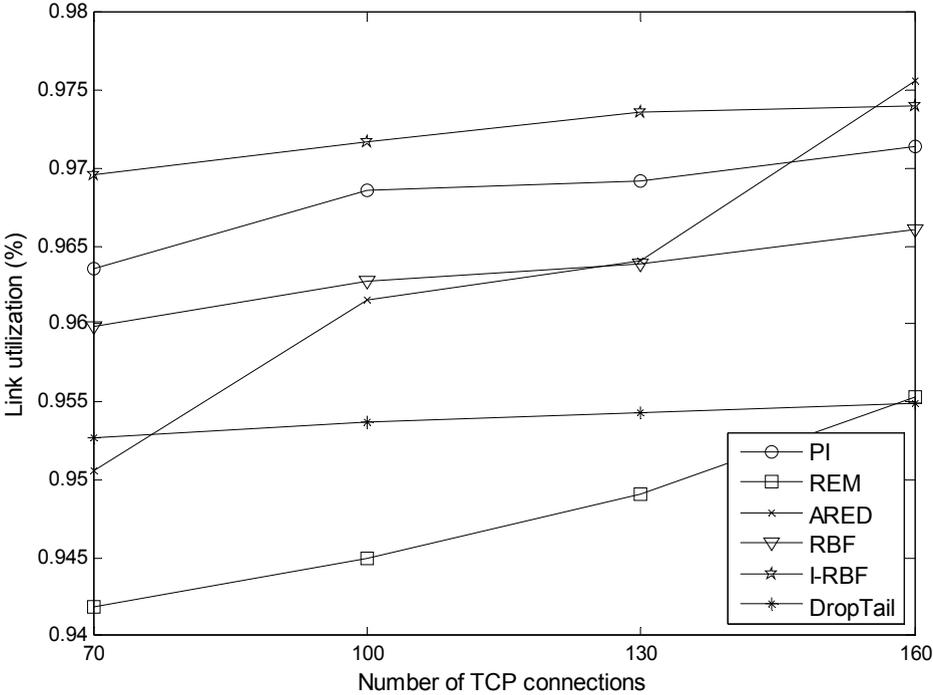

**Fig. 5.** Link utilization against number of connections of proposed controllers compared to other queue control schemes

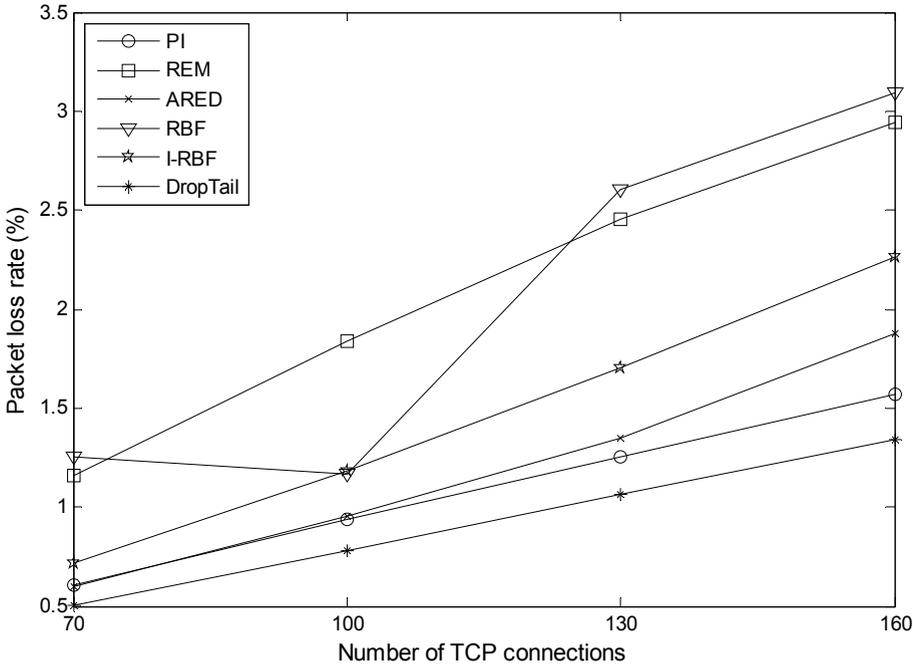

**Fig. 6.** Packet loss rate against number of connections of proposed controllers compared to other queue control schemes

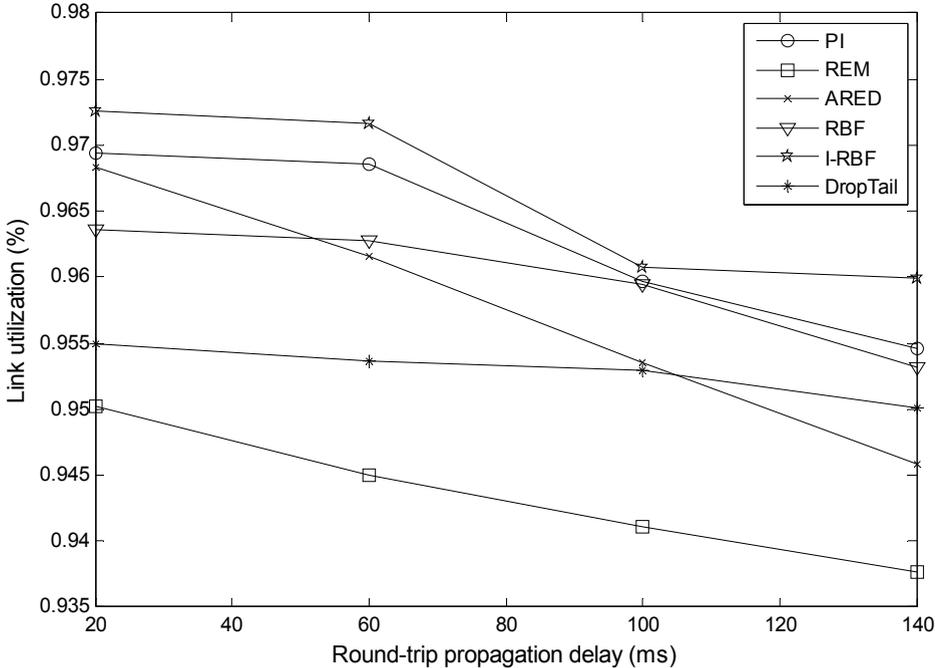

**Fig. 7.** Link utilization against propagation delay of proposed controllers compared to other queue control schemes

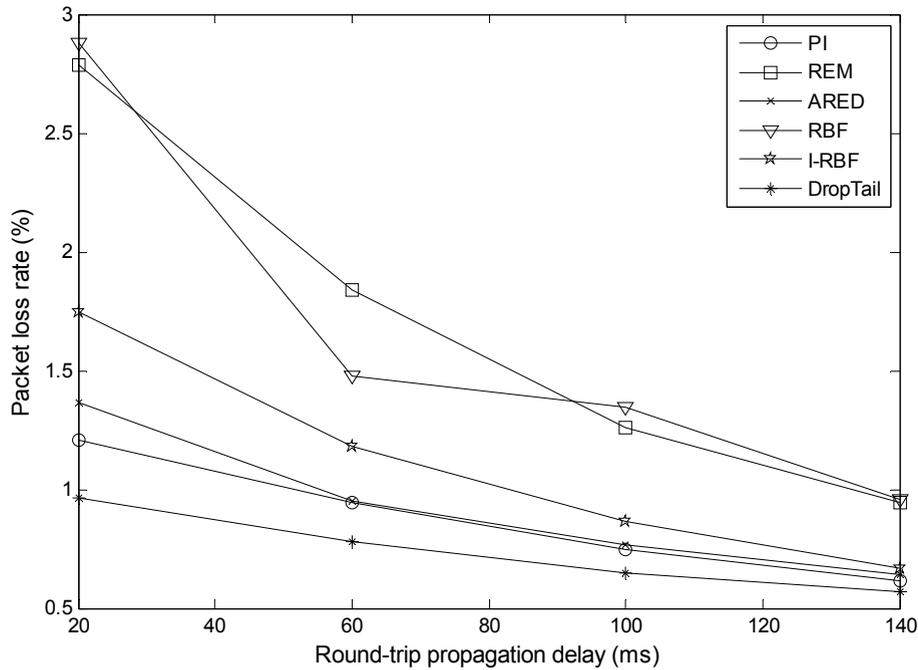

**Fig. 8.** Packet loss rate against propagation delay of proposed controllers compared to other queue control schemes

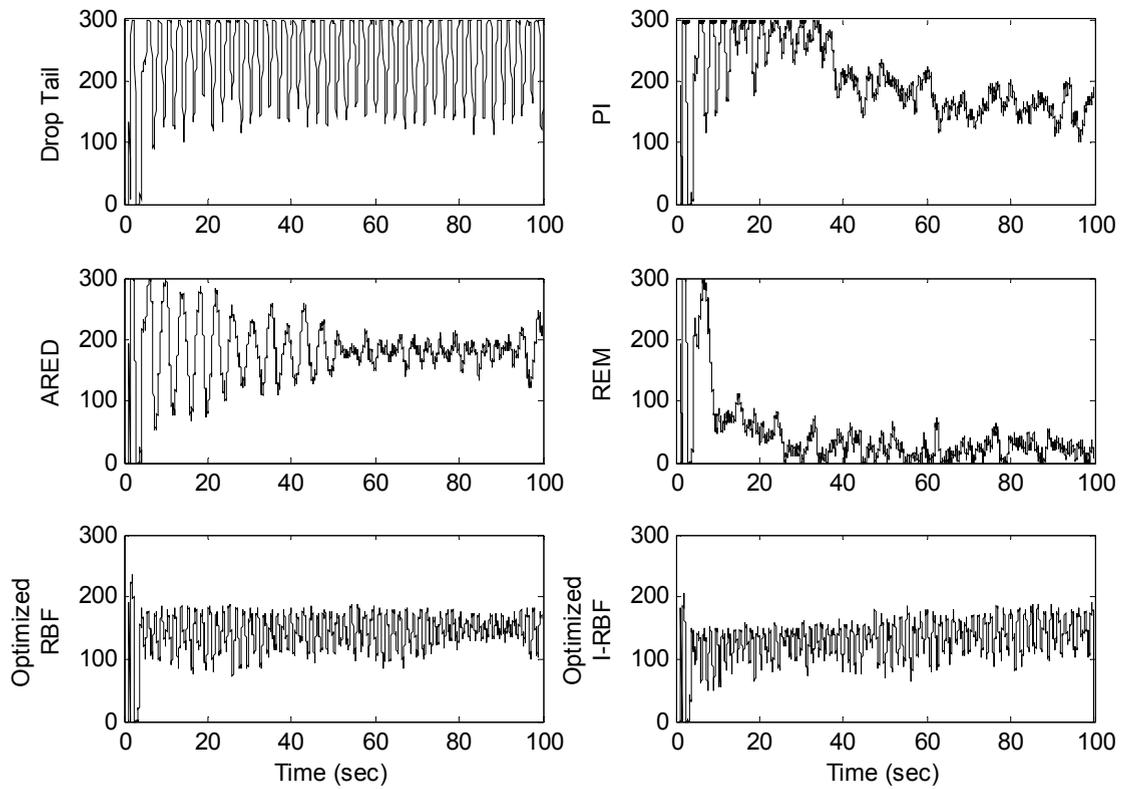

**Fig. 9.** Queue size of the proposed schemes in Scenario 1 compared to other queue control schemes

*Scenario 1: Constant TCP-Reno connections*- Figure 9 shows the simulation results in this scenario when using GA-PSO-optimized RBF-based controllers.

As can be seen, the optimized RBF and optimized I-RBF are converged to the desired queue size more rapidly as compared to other mentioned queue control schemes. Although, the responses of PI and ARED can achieve and maintain the queue length around the desired value, too inactive and serious overshoots occurred. Since there is no parameter of the target queue length in the REM control algorithm, so it cannot maintain the desired queue length.

*Scenario 2: Dynamic traffic load*- In this scenario, the performance of different active queue control schemes considering dynamic network traffic are evaluated. At the beginning, $t = 0$, 100 TCP-Reno connections have been established. Then, at $t = 30$, 30 more TCP-Reno connections begin transmission and remained inactive until $t = 60$. Additionally, 30 TCP connections departed at the same time, so there is only 70 active connections till $t = 80$, which 100 connections would transmitt till the end of the simulation period. Fig. 10 shows the corresponding queue evolutions obtained for the different queue control schemes. It can be seen that Drop Tail, PI, ARED and REM controllers are not robust with respect to variations in the load. On the other hand, the proposed GA-PSO-optimized RBF and I-RBF controllers are robust to variations in the number of active connections.

*Scenario 3: Short and long propagation delays*- The robustness of the proposed methods against variations of the RTT is evaluated in this scenario. At first, we assume that the links between senders and $R_1$ and also between $R_2$ and receivers are characterized by 10 Mbps-bandwidth and a short inherent propagation delay of 2 msec. There is a bandwidth of 10 Mbps and an inherent propagation delay of 10 msec between $R_1$ and $R_2$. The responses of the queue length obtained for different queue control schemes are shown in Fig. 11.

The effect of a long inherent propagation delay is also considered, where the TCP sources and sinks are linked respectively to $R_1$ and $R_2$ with 10 Mbps bandwidth and 20 msec propagation delay. Also, a propagation delay of 140 msec is chosen between $R_1$ and $R_2$. The values of queue length obtained with different queue control schemes are shown in Fig. 12. As can be seen, ARED has a steady-state error with a short propagation delay and has serious oscillation when given a long propagation delay. Again, Drop Tail is not robust and shows periodic behavior. Even though PI has no steady-state error in the network with long propagation delay, but the transient responses are too sluggish. The proposed GA-PSO-optimized RBF controller has small steady-state error, but the optimized I-RBF achieved the desired queue length in a reasonable transient response time.

## 7. CONCLUSION

This paper has presented a hybrid evolutionary-swarm-neural model for adaptive queue controller to enhance TCP congestion control in communication networks. In this way, GA has been used to determine the optimum number of RBFs in neural model and PSO algorithm has been used to tune the weights of mentioned controller.

In order to improve the robustness of proposed method an improved RBF controller has been introduced, too. A deep analysis of stability and performance of the proposed schemes has been carried out through simulations using ns-2 tool through different scenarios. The simulation studies have demonstrated that the proposed GA-PSO-optimized I-RBF controller outperforms the Drop Tail, PI, REM and ARED queue control policies under various operating conditions. It achieves both good queue regulation and high link utilization. It has also shown that the proposed controller has fast response and is robust to high load variations and disturbance rejection in steady-state behavior. In addition, the link utilization of proposed controller is higher than mentioned simulated controllers in this study, with a low packet loss.

## 8. ACKNOWLEDGEMENT

This work is supported by Islamic Azad University-South Tehran Branch under a research project entitled as "Design and Simulation of Optimal Neural Controllers for Active Queue Management in TCP Communication Networks".

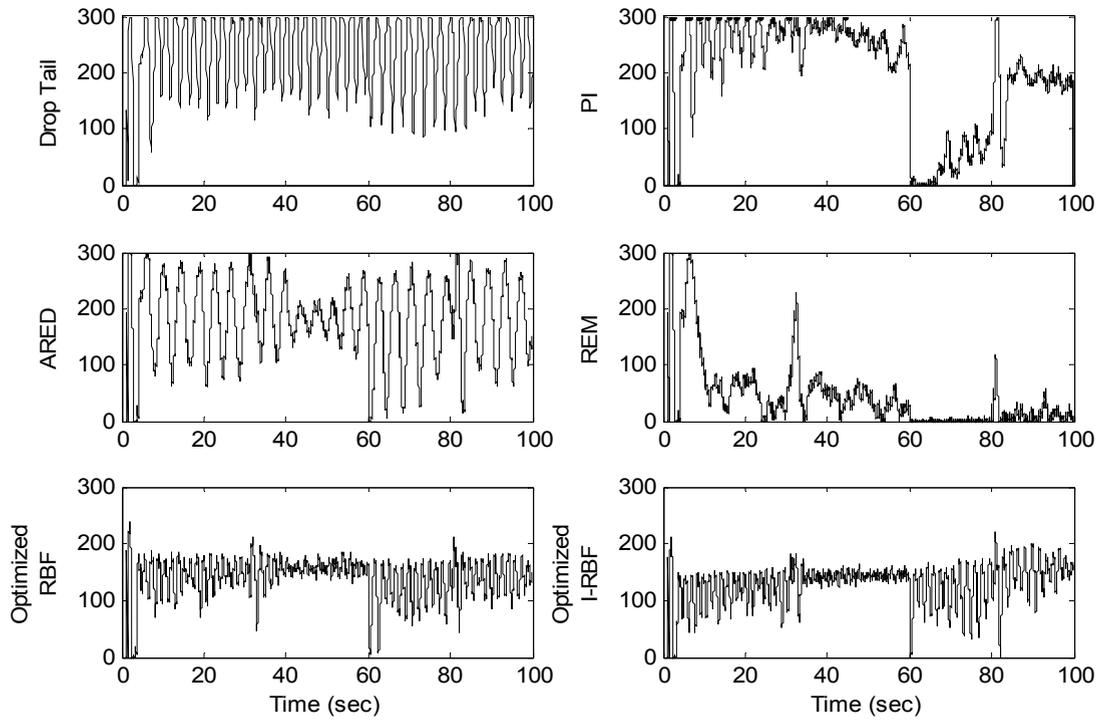

**Fig. 10.** Queue size of the proposed schemes in Scenario 2 compared to other queue control schemes

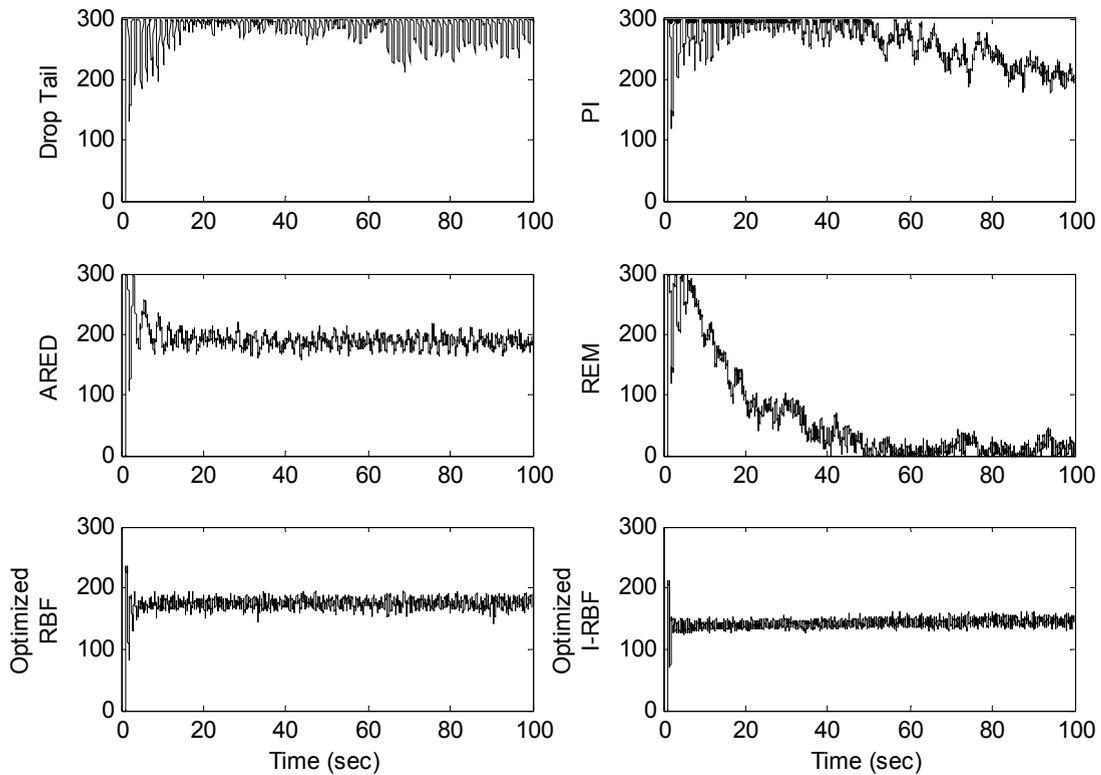

**Fig. 11.** Queue size (in packets) of the proposed schemes in short propagation delay times compared to other queue control schemes

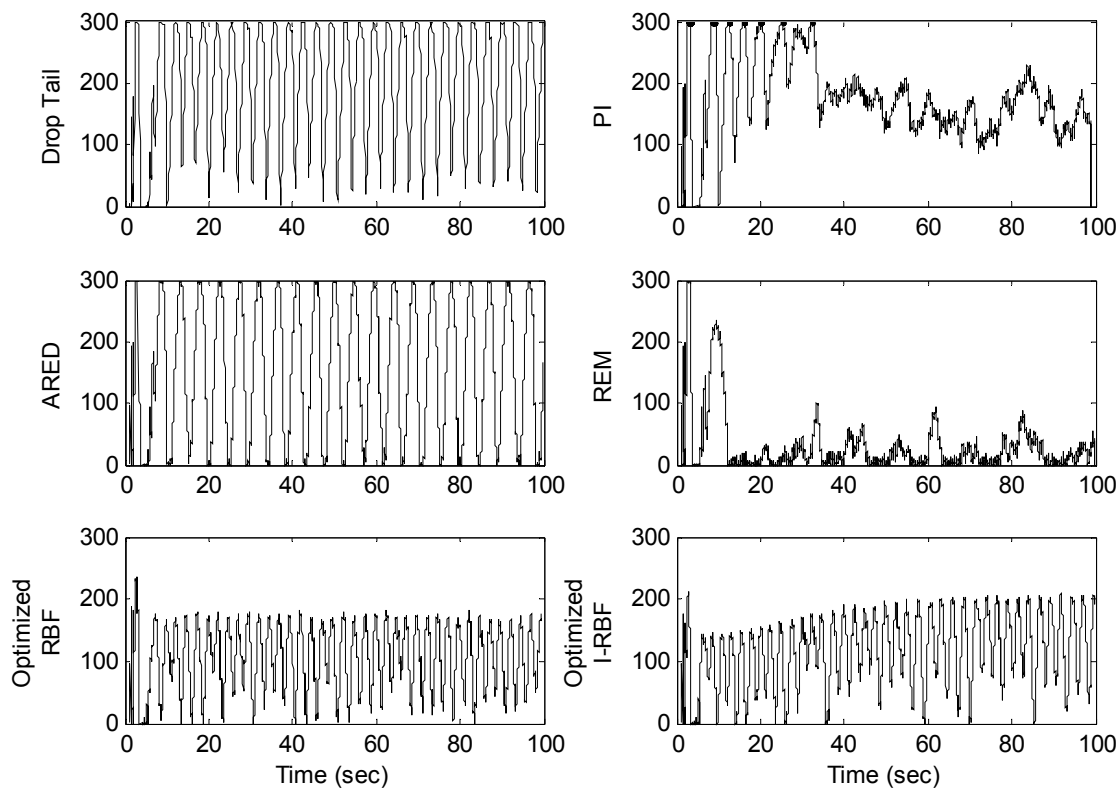

**Fig. 12.** Queue size (in packets) of the proposed schemes in long propagation delay times compared to other queue control schemes